\documentclass[letterpaper, 10 pt, conference]{ieeeconf}  

\IEEEoverridecommandlockouts                              

\usepackage{xcolor}
\usepackage{hyperref}
\usepackage{url}
\usepackage{graphicx}
\usepackage{authblk}

\title{\LARGE \bf DEVI: Open-source Human-Robot Interface for \\Interactive Receptionist Systems}

\author{Ramesha Karunasena${}^{1\dagger}$, Piumi Sandarenu${}^{2}$, Madushi Pinto${}^{3}$, Achala Athukorala${}^{4}$, \\ Ranga Rodrigo${}^{5}$ and Peshala Jayasekara${}^{6}$}

\affil{Department of Electronic and Telecommunication Engineering, University of Moratuwa, Sri Lanka\\
\href{mailto:me@somewhere.com}{\{140292g${}^{1}$, 140557b${}^{2}$, 140466u${}^{3}$, 130048g${}^{4}$, ranga${}^{5}$, peshala${}^{6}$\}@uom.lk}}

\begin{document}

\maketitle
\thispagestyle{empty}
\pagestyle{empty}

\begin{abstract}

Humanoid robots that act as human-robot interfaces equipped with social skills can assist people in many of their daily activities. 
Receptionist robots are one such application where social skills and appearance are of utmost importance. 
Many existing robot receptionist systems suffer from high cost and they do not disclose internal architectures for further development for robot researchers.
Moreover, there does not exist customizable open-source robot receptionist frameworks to be deployed for any given application.
In this paper we present an open-source robot receptionist intelligence core---``DEVI"(means 'lady' in Sinhala), that provides researchers with ease of creating customized robot receptionists according to the requirements (cost, external appearance, and required processing power). 
Moreover, this paper also presents details on a prototype implementation of a physical robot using the DEVI system. The robot can give directional guidance with physical gestures, answer basic queries using a speech recognition and synthesis system, recognize and greet known people using face recognition and register new people in its database, using a self-learning neural network. 
Experiments conducted with DEVI show the effectiveness of the proposed system.

\end{abstract}

\begin{keywords}
Social robot system, human-machine interface, humanoid robot, intelligent system, robot receptionist
\end{keywords}

\section{INTRODUCTION}

The duty of a receptionist is to assist people in a friendly and pleasing manner. 
Nevertheless, human receptionists are unable to maintain their composure, focus and efficiency at a consistent level at all times. 
As a result, the quality of service provided by human receptionists can deteriorate due to many factors including service time and the number of service requests. 
Moreover, the job of a receptionist can sometimes be dull and monotonous, which leads them to fall asleep on their desks. 
Furthermore, many additional factors such as down-times (breaks, vacations) need to be considered when employing human receptionists. 
To address the above issues and to improve the quality of service provided by receptionists, we can introduce robotic receptionist systems. Robot receptionists can introduce excitement and memorable experience for the interacting humans.

A robot receptionist can be considered as a Socially Assistive Robot (SAR), since its objective is to assist human users through human interaction~\cite{feil2005defining}, or it can be considered as a Socially Interactive Robot (SIR)---a robot equipped with social interaction skills, as defined by Fong et al.~\cite{fong2003survey}. 
Both SAR and SIR belong to the social robotics category. 
One of the key goals of social robotics is to implement an intelligent system with functional social interactive capabilities.
Extensive research is being carried out in this domain and sophisticated humanoid robotic systems with advanced capabilities such as Sophia~\cite{weller2017meet} can be identified as the state-of-the-art.
However, almost all existing innovations are high-priced and are barely suitable for public or commercial use. 
Furthermore, it is also time consuming to develop custom solutions for each specific application, since there does not exist any open-source or commercial intelligence cores/platforms for developing receptionist robots, to the best of our knowledge.\par

Consequently, the main objective of this work is to develop a low-cost, user-friendly receptionist robot with an open-source intelligence core, which can assist human users in a friendly manner. 
Our solution: DEVI is the first ever human-like robot receptionist in Sri Lanka and the first open-source intelligence core specifically aimed at robot receptionists. Our robot has the following functionalities and features.\par

\begin{itemize}
  \item Face identification (of known people)
  \item Face memorization and re-identification (of unknown/ new people)
  \item Speech recognition and synthesis
  \item Conversational chatbot
  \item Direction guidance with physical hand gestures
  \item 180-degree person proximity sensing and interaction with a rotatable head unit
\end{itemize}

The rest of the paper is organized as follows: section~\ref{RELATED WORK} describes related work; section~\ref{SYSTEM ARCHITECTURE AND DESIGN} explains the overview and design criteria of the proposed system; achieved results of the proposed system are given in section~\ref{RESULTS}; finally, in section~\ref{CONCLUSION}, important conclusions on the proposed robot receptionist system are made, while suggesting possible future improvements. \par

\section{RELATED WORK}
\label{RELATED WORK}



Pepper~\cite{lafaye2014linear} is a personal assistant robot introduced in 2014, which has been adopted by many users to replace human receptionists. Pepper is popular in countries like Japan, UK and Belgium and costs around \$22,000~\cite{pepperprice} for purchase. It is capable of detecting human emotions, performing hand, neck and several body motions. Nevertheless, Pepper has a more machine-like appearance and is considered to be expensive to be adopted for small and medium scale establishments. The intelligence core of Pepper is not disclosed to developers to create custom solutions for different application scenarios. \par

Sofia~\cite{weller2017meet} is a revolutionary humanoid robot introduced by Hanson Robotics as a social humanoid robot in 2015 with advanced potential to closely mimic a human. 
It has the capability of displaying more than 50 expressions and it consists of an offline AI core. 
Experts who have reviewed the open-source code of the Sophia robot state that Sophia is best categorized as a chatbot with a face~\cite{gershgorn2017inside}.

Aiko Chihira is a humanoid robot assistant deployed to work as a department store receptionist~\cite{aiko}. She is capable of performing acts such as greeting and providing information to the customers. She is also designed with advanced technology to mimic human appearance and facial expressions.
Nadine~\cite{thalmann2017nadine} can be identified as another social humanoid robot, which is modeled after Professor Nadia Magnenat Thalmann. 
Nadine is currently deployed as a receptionist. \par

The proposed robot receptionist system, DEVI, is motivated and inspired by the aforementioned receptionist systems of varying complexities and cost. The main objective of this proposed work is to design and develop an open-source, low-cost, user-friendly, customizable receptionist intelligence with a human-like appearance, mainly to be utilized at small and medium scale establishments. It is also intended as the initial step of a research platform that can capture human interactions for future study.

\section{SYSTEM ARCHITECTURE AND DESIGN}
\label{SYSTEM ARCHITECTURE AND DESIGN}

\begin{figure}[tb]
\begin{center}
\includegraphics[width=\columnwidth]{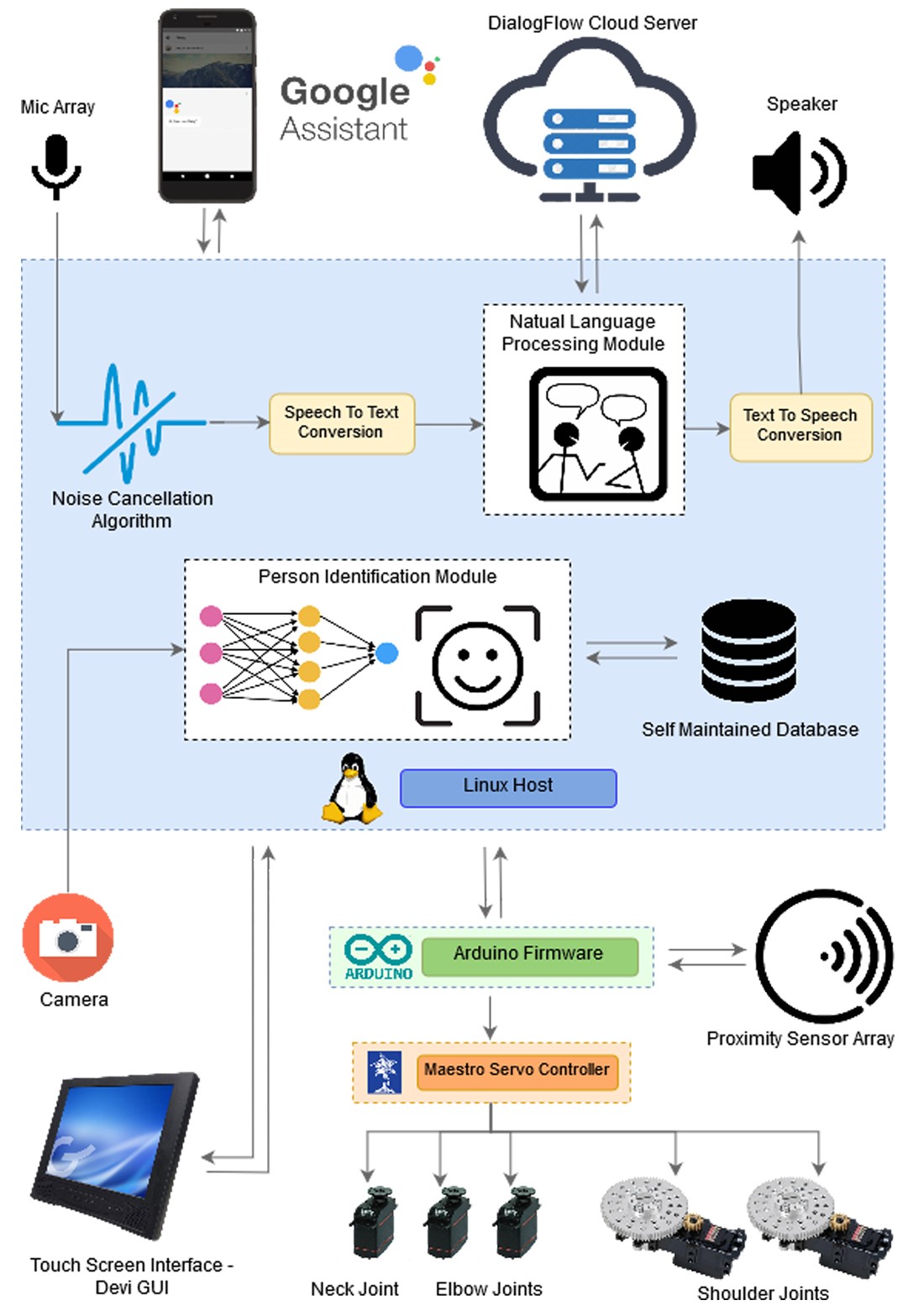}
\caption{\label{architecture} System architecture of DEVI}
\end{center}
\end{figure}

\begin{figure}[tb]
\begin{center}
\includegraphics[width=\columnwidth]{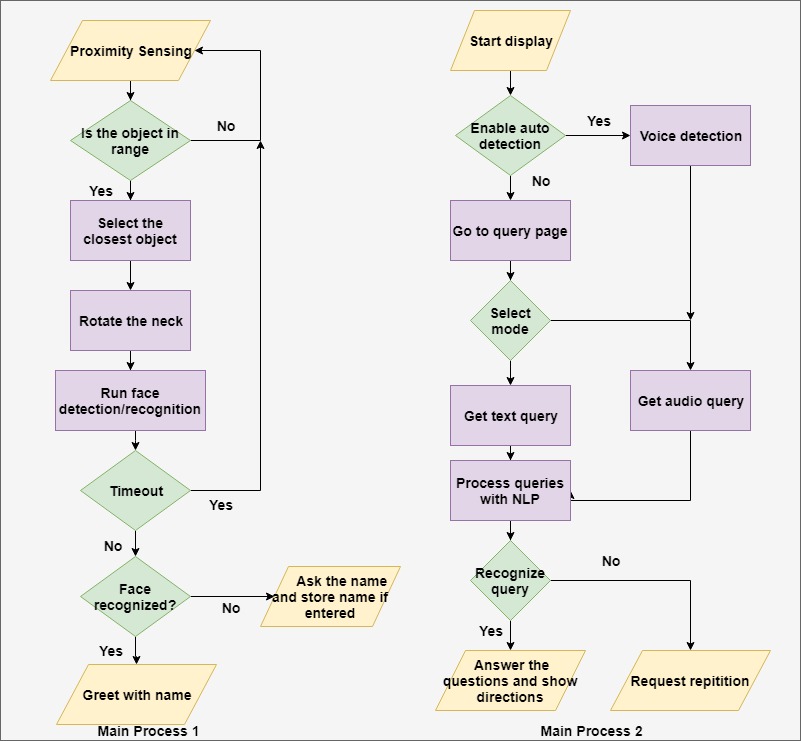}
\caption{\label{systemflow} DEVI interactive processes: (a) Main Process 1- Person detection and identification  (b) Main Process 2- User query services}
\end{center}
\end{figure}


The overall system architecture of the DEVI robot receptionist is shown in Figure~\ref{architecture}. DEVI system consists of a loosely coupled modular architecture, which isolates modular failures without causing a total system crash. Furthermore, this modular architecture also facilitates the addition of new features to the design easily. DEVI's physical structure is designed using a mannequin, to ensure a human-like appearance. DEVI's system flow diagram can be explained in two main processes as shown in Figure~\ref{systemflow}.

A novel feature, which highlights DEVI from existing robot receptionists, is the proximity sensing necklace. This proximity detection unit generates a proximity map with a 180-degree field of view of the robot's surrounding. Using the proximity map, DEVI is able to accurately turn its head to interact with people approaching from different directions. 

The proximity map also helps in controlling the execution of facial recognition process to reduce processor utilization. 
The facial recognition process is triggered only when a person is in close proximity to the robot. 
DEVI analyses three possible scenarios for human interaction initiation: known-person identification, unknown person detection and a false detection. 
In a known-person identification scenario, DEVI will greet the person using the speech synthesis and limb actuator system. 
In an unknown-person detection scenario, DEVI will ask for the name of the person and store the person's face in a dynamic database with the person's consent, so that it can detect his/her face in future encounters. 

In a false detection scenario, DEVI will return to the initial idle state. False positives are identified when a face is not detected within a predefined timeout.
Currently, DEVI only supports the English language for speech recognition and synthesis. A touch display is provided with the robot, as an auxiliary interface, to assist people who may fail to speak in the standard English accent. DEVI system design is explained in the following sections. \par

\subsection{Hardware Layer}
The proximity detection unit detects people within a specified maximum range, creates a proximity map and triggers the face detection system accordingly. 
The neck-motion control unit is responsible for controlling the direction of the robot's head by turning the neck, based on the detection angle of the user (the relative orientation of the user, with respect to the robot). 
The detection angle is extracted from the proximity map, which is generated by the proximity detection unit.
The limb actuator control system controls the hands of the robot to give directional guidance using physical hand gestures. The touch display, camera, mic array and the speakers are connected to the Linux host computer, while the proximity detection unit and the motor controller for limb and neck actuators are interfaced with the robot's main controller, implemented using an ATmega2560 microcontroller. The main controller and the Linux host maintains a serial communication link, using a custom packet structure.

\subsubsection{Person Proximity Detection Unit}
This unit comprises five VL53L0X time-of-flight (TOF) sensors~\cite{c7}, each with a field of view of 25-degrees and they are placed with a gap of 45-degrees between two sensors; the circular arrangement appears as a necklace on DEVI. The unit can sense within a radius of 110\,cm with a 180-degree field of view. 
Whenever users appear in the proximity map, the proximity detection unit will register all these events using a First-In, First-Out (FIFO) data structure.
In the presence of many users, the ATmega2560 controller has the ability of prioritizing regions in the proximity map generated by the proximity detection unit.
Each time a conversation is finished, the main interface will dequeue the next element of the FIFO queue, which corresponds to the region where the next user has arrived, and thereby the robot will service the corresponding person.

\subsubsection{Limb Actuator Control System}

\begin{figure}[t]
\begin{center}
\includegraphics[width=\columnwidth]{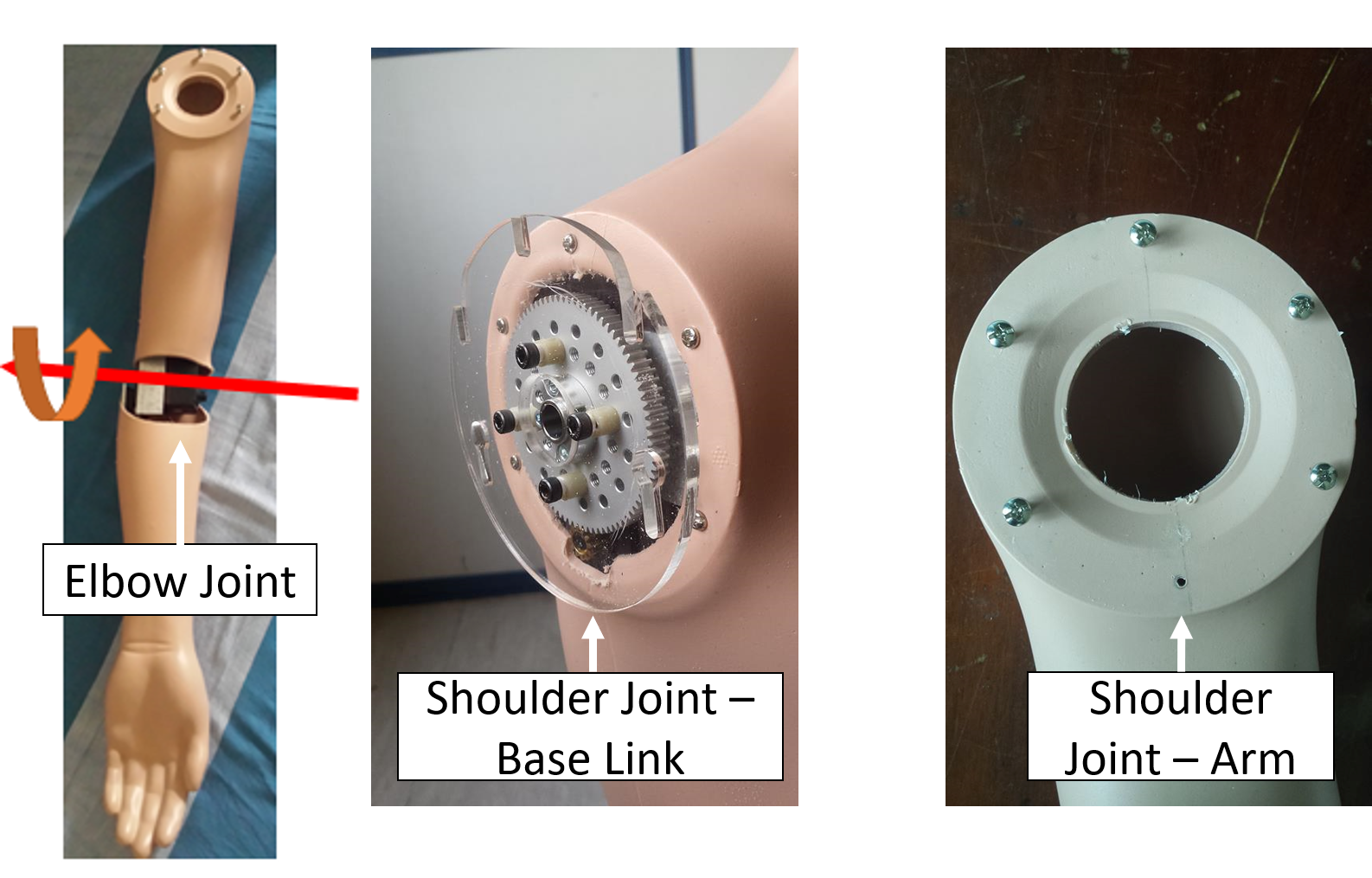}
\caption{\label{arms} Elbow and shoulder joints of DEVI}
\end{center}
\end{figure}

This system is responsible for the control of two degrees of freedom in each hand: shoulder joint and elbow joint. 
It comprises two CM-785HB (Hitec) servo motors~\cite{c8} with external gear reducers for shoulders and two SpringRC SR811 servo motors~\cite{c9} for elbows. The elbow and shoulder joints are shown in Figure~\ref{arms}. All the motors in the system are controlled by a Micro Maestro 6-Channel USB Servo Controller~\cite{c10}, which is interfaced with the ATmega2560 microcontroller that communicates with the Linux host system through serial communication.

\subsubsection{Neck-motion Control Unit}
The neck motion is achieved by a single degree of freedom at the neck joint. A TowerPro MG945 servo motor~\cite{c19} is fixed inside the head of the robot to rotate the head. It enables DEVI to rotate its head towards the user, based on the input from the person proximity detection unit. This unit only rotates the robot's head, keeping the orientation of the proximity detection unit fixed.

\subsection{Robot Intelligence Core}
DEVI's intelligence core comprises three main units: face recognition unit, chatbot and the human-machine interface (HMI). 
Deep learning based face recognition system allows DEVI to recognize people who have been previously encountered. The chatbot facilitates the acquisition of the identities (name, face image) of unknown people to maintain a dynamic database. Later, DEVI can retrain her face recognition neural network periodically during idle times, using this dynamic database. 
The speech recognition, synthesis and natural language processing (NLP) components within the chatbot system assist DEVI to further enhance her social skills and answer user queries. The HMI comprises two main units: the auxiliary touch display and the integrated Google assistant.

\subsubsection{Face Recognition System with Dynamic Database}

\begin{figure}[tb]
\begin{center}
\includegraphics[width=\columnwidth]{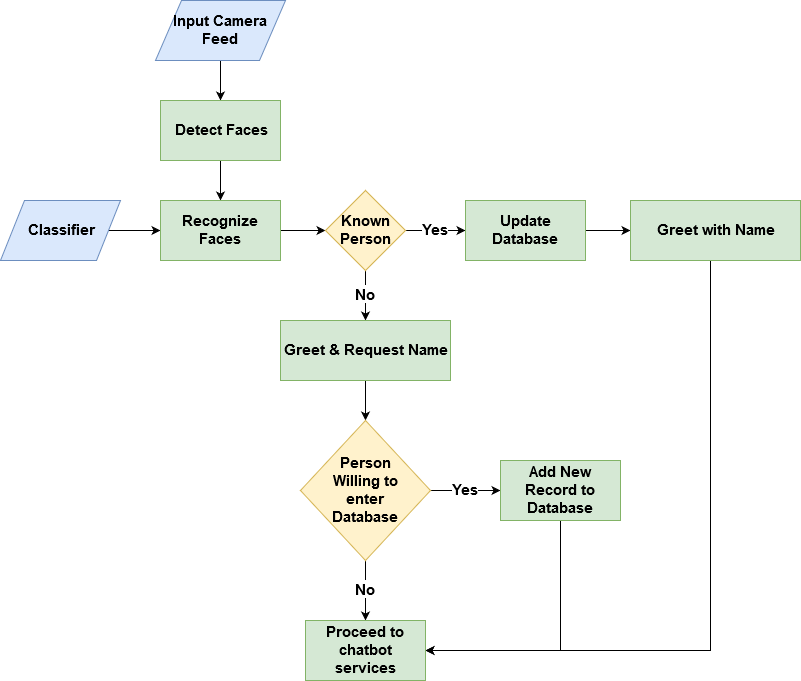}
\caption{\label{face} Flow diagram of the face recognition system}
\end{center}
\end{figure}

The main execution loop of the face detection and recognition design include the following sequence of steps.

\begin{itemize}
  \item Receive and sample input video stream 
  \item Detect the presence of one or more faces 
  \item Recognize faces, if the faces are found within database
  \item Add to the database if faces are not recognized  
\end{itemize}

Face detection and face recognition tasks are activated only when the proximity sensor readings indicate the presence of an object. 
In this event, the live video feed from the camera is fed to the face recognition system. In the absence of a face being detected by the face detection algorithm, the robot will move back to the initial idle state.\par

If a face is detected, the robot will attempt to recognize the person, and if the person is recognized, DEVI will proceed to greet the person by his/her name. However, if the person is not recognized, he/she will be greeted and requested to enter the name, in order to be registered in the robot's dynamic database of users. 
When the user is willing to get registered with the system, facial images of the user will be saved in the database along with the name. 
However, if the user has not consented to be registered, the robot will directly service the user's queries. 
Figure~\ref{face} illustrates the flow of the face recognition process. The robot has a local database implemented using `mongoDB'~\cite{c11} database management software, which stores the name of each person, number of instances the person has been recognized by the robot and the last recognized date and time.\par

The face detection and recognition unit is implemented using a deep residual network implementation (Resnet-29)~\cite{c12,c13} provided by `dlib'~\cite{c14} and a k-nearest neighbors (KNN) classifier. Face detection is carried out using a Histogram of Gradient (HoG) based feature detector available in `dlib'. 
This system is capable of detecting faces which are of 40$\times$40 pixel size or higher. In comparison with the convolutional neural network (CNN) based feature detector, this approach is much faster. 
The KNN classifier is used to compare several face poses of the same person. 
We utilize 10 images of each person accumulated over time when the person approaches the robot in several occasions. This threshold can be changed, and more faces would result in higher accuracy but longer training cycles to generate the KNN model. The face recognition algorithm executes in less than 67\,ms. In order to improve the recognition accuracy, multiple sample frames are fed through the algorithm iteratively.\par

\subsubsection{Speech Recognition, Synthesis and NLP System}

\begin{figure}[tb]
\begin{center}
\includegraphics[width=\columnwidth]{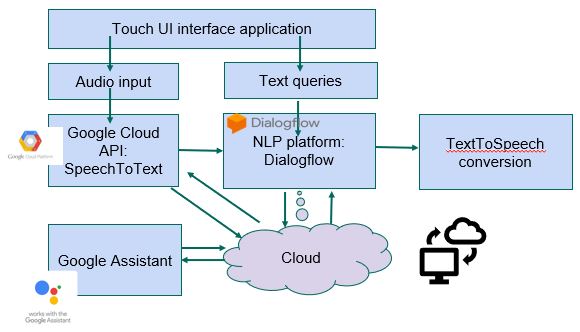}
\caption{\label{speech} Overview of the DEVI chatbot}
\end{center}
\end{figure}
Speech recognition, synthesis and NLP systems are used to greet people, get to know their identities, maintain a conversation by responding to basic questions and etiquettes, and give directional guidance for known places. The NLP also features intent mapping, where the user can deliver the same query in different wordings; formal, informal or partially complete.\par

The speech recognition module has 2 modes of operation: Auto-detection Enabled and Disabled (Figure~\ref{systemflow}). 
If auto-detection is enabled, voice clips exceeding a specified amplitude threshold will be extracted and processed using ReSpeaker Mic Array V2.0 \cite{mic}. If autodetection is disabled, then the user can choose either audio mode or text mode. 
In audio mode, the user can send his/her queries in audio clips of 5 seconds, which will be converted to text using the speech-to-text converter, and then will be passed to the NLP module. 
The output from the NLP module will be given to the text-to-speech converter and to the speaker embedded within the head of DEVI robot. If the text mode is selected, the process will skip the speech-to-text conversion and the original text query will be directly sent to the NLP module. Figure~\ref{speech} illustrates the inner workings of the DEVI chatbot.

The audio queries are processed and converted to text queries using Google Cloud API. 
The text queries are directed to the agent hosted in the Dialogflow~\cite{c15} cloud-based platform, where NLP is carried out for the trained data. 
The output from the NLP module is subjected to the text-to-speech conversion, in which Pyttsx~\cite{manaswi2018speech} Python package is utilized.
The intermediate text output from the NLP module will be displayed in the touch display for user convenience. 
For this specific application, DEVI's NLP model has been re-trained using an existing generic NLP model, also to give directional guidance to places inside the Department of electronic and Telecommunication Engineering of University of Moratuwa. However, the user can use their own Dialogflow model re-trained for their specifications and use the core platform of DEVI easily, due to the customizability of the proposed core platform.\par

\subsubsection{Human-Machine Interface}

\begin{figure}[tb]
\begin{center}
\includegraphics[width=\columnwidth]{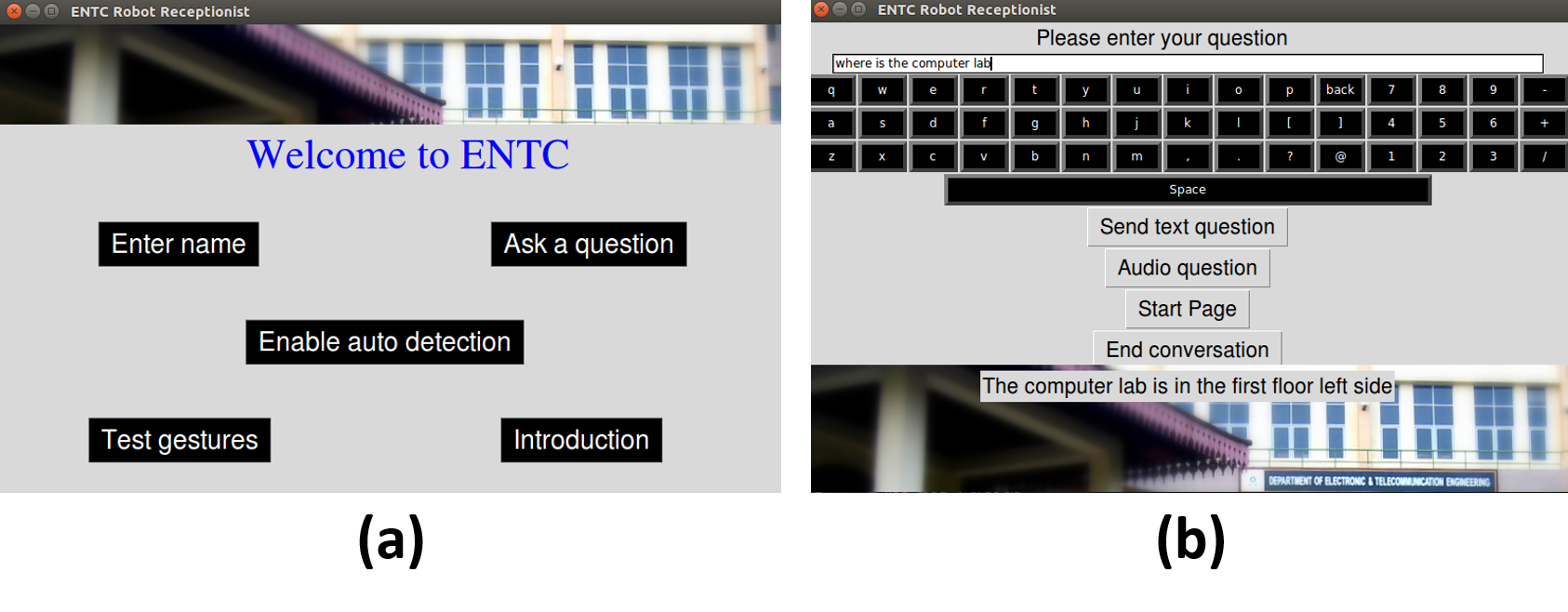}
\caption{\label{welcome}(a)~The main page of DEVI GUI (b)~Page to access DEVI chatbot in audio or text modes}
\end{center}
\end{figure}


The intelligence core's HMI consists of a graphical user interface (GUI) and an integration to the Google Assistant. \par 

DEVI's display application is designed using Python Tkinter~\cite{lundh1999introduction} for the auxiliary touch display (GUI). As a socially assistive robot, DEVI also serves people with speech disorders by interacting through the provided touch display. The display application also acts as an alternative for speech interaction when the user is unable to communicate in a standard English accent with the robot. 
The GUI application has the following main controls: enter user name, access chatbot in audio/text mode, activate audio autodetection, test the gestures of DEVI and give a self-introduction of DEVI. Functions such as selecting modes either text or audio query are operated using the GUI application. Figure~\ref{welcome} shows two screenshots of the DEVI GUI. \par
Apart from the GUI, we also have integrated the speech recognition module with Google assistant~\cite{forrest2017essential} in order to collect robot data from an offsite location. 
The chatbot is integrated with Google assistant and is capable of communicating through any Android smart phone with authorized access. Hence, we utilize 3 methods to interact with the robot: Google assistant, speech queries and text queries from display application.\par

\section{EXPERIMENTS AND RESULTS}
\label{RESULTS}

\begin{figure}[t]
\begin{center}
\includegraphics[width=\columnwidth]{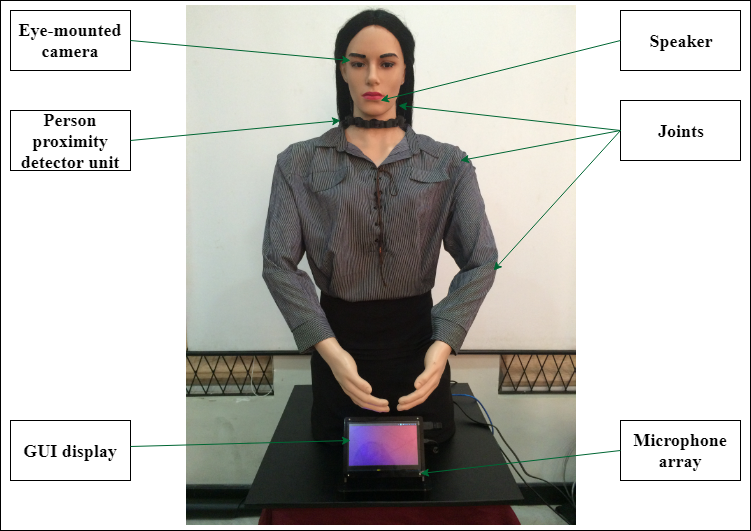}
\caption{\label{DEVI} DEVI, the robot receptionist}
\end{center}
\end{figure}

This section describes the experiments and corresponding results obtained for the proposed system. Figure~\ref{DEVI} and accompanying videos~\cite{youtube} \cite{video} show the actual implementation of DEVI, the robot receptionist. The code base of DEVI is openly available in the given GitHub repository~\cite{devigit}.\par

\subsection{General Specifications}

\begin{itemize}
  \item Robot weight: 9.9\,kg (including the platform)
  \item Robot height: 110\,cm
  \item Max current consumption: 1.32\,A
  \item Idle current consumption: 0.18\,A
  \item Robot operating voltage: 24\,V
\end{itemize}

\subsection{Actuator Control System}

The results of the actuator control system are given in Table~\ref{table:actuator-control-sys}.

\begin{table}[b]
\caption{Results of the actuator control system}
\label{table:actuator-control-sys}
\begin{center}
\begin{tabular}{|c|c|c|}\hline
Joint & Rotating angle & Speed \\
 & (degrees) & (degrees/second) \\\hline
Shoulders (2) & 180 & 36.5 \\\hline 
Elbows (2) & 110 & 69 \\\hline
Head & 170 & 23.3 \\\hline
\end{tabular}
\end{center}
\end{table}

\begin{table}[tb]
\caption{Statistics of sample TOF data sets}
\label{table:tof}
\begin{center}
\begin{tabular}{|c|c|c|}\hline
Smoothing & Mean & Standard \\
Factor & (mm) &  Deviation (mm) \\\hline
1 & 1014.47 & 5.94 \\\hline 
0.1 & 1015.61 & 2.07 \\\hline
0.3 & 1016.85 & 3.34 \\\hline
0.5 & 1020.71 & 3.87 \\\hline
0.7 & 1019.95 & 4.41 \\\hline
0.9 & 1016.70 & 5.69 \\\hline
\end{tabular}
\end{center}
\end{table}

\subsection{Person Proximity Detection Unit}

A simple linear recursive exponential filter is used to smooth out the noisy distance measurements obtained from the TOF sensors. The comparison of results obtained for a data set of 500 TOF readings at a distance of 1\,m with a varying smoothing factor is given in Table~\ref{table:tof}. Thus, the optimum smoothing factor of 0.1 with the lowest standard deviation was selected.\par

\subsection{Face Recognition System}
This section presents the implementation and test results of the face recognition system of the DEVI robot.\par
\begin{figure}[tb]
\begin{center}
\includegraphics[width=\columnwidth]{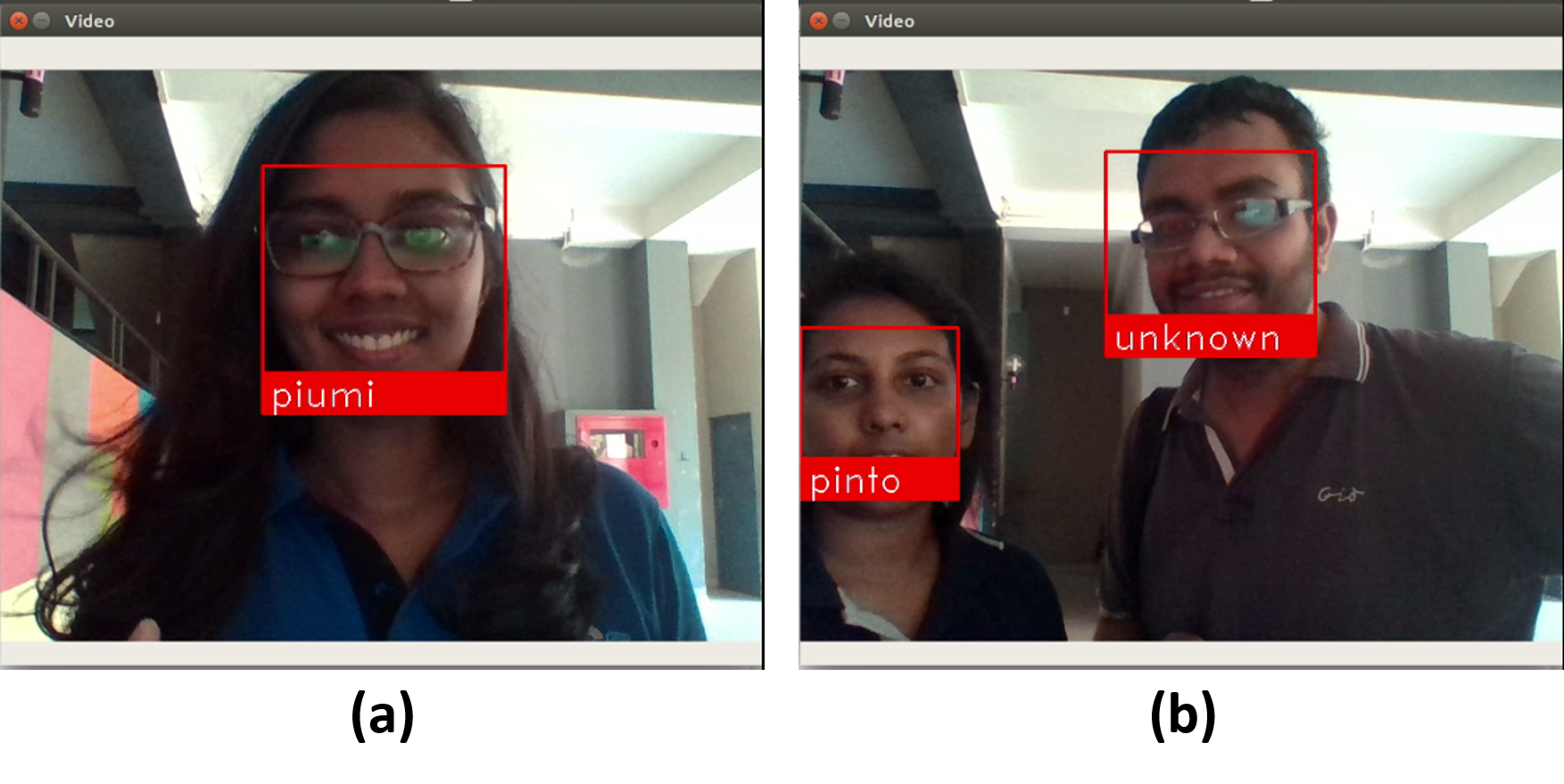}
\caption{\label{facescreenshots} Recognition instances of users (a)~Known user (b)~Unknown user}
\end{center}
\end{figure}

Figure~\ref{facescreenshots} shows few recognition instances from the face recognition system.
Images were obtained using a camera with a 1.3\,Megapixel image sensor (1280$\times$1024 line resolution) and a 30-frames per second frame rate. The video feed is sampled by the face recognition module at a frequency of 15\,Hz. The measured overall performance accuracy of the face recognition system is 97.7\% against faces from the people available in ‘Faces in the Wild’ dataset~\cite{c20}. The performance results of face detection and feature extraction are presented in Table~\ref{table:faceperformance}.
\par

\begin{table}[tb]
\caption{Performance of face detection and feature extraction steps}
\label{table:faceperformance}
\begin{center}
\begin{tabular}{|p{5cm}|p{1.7cm}|}\hline
Performance measure & Time (s) \\\hline
Average time for face detection & 0.02 \\\hline
Average time for feature extraction using the Resnet-29 model & 0.37 \\\hline
\end{tabular}
\end{center}
\end{table}

\subsection{Speech recognition, synthesis and NLP}

\begin{table}[t]
\caption{Results of speech recognition, NLP and synthesis}
\label{table:NLP}
\begin{center}
\begin{tabular}{|p{3cm}|p{0.7cm}|p{3cm}|}\hline
Conditions & Accu & Comments\\
&racy&\\\hline
Responses to different variations of intents & 91.7\% & Text queries \\\hline 
Responses to different users & 77\% & Speech queries with different users in an environment with less noise \\\hline
Responses of the chatbot in a noisy environment & ~50\% & Accuracy varies in a large range \\\hline
NLP server response time & \textless2s & Depends on network speed \\\hline
Audio response time & ~6.5s & Depends on network speed \\\hline
\end{tabular}
\end{center}
\end{table}

The results of speech recognition, synthesis and NLP are tabulated in Table~\ref{table:NLP}. When operated in text mode, the NLP module has an accuracy of 91.7\% and the accuracy can be improved with further training. However, when operated in audio mode, accuracy varies in a wide range from 77\% to ~50\%, due to the effect of ambient noise. In a noisy environment, it is difficult for the speech-to-text converter to transcribe the audio query to text. Using improved noise-cancellation algorithms or dedicated noise-cancelling hardware for audio processing can minimize this issue.\par

\section{CONCLUSION}
\label{CONCLUSION}

\subsection{Summary}
DEVI is a humanoid robot receptionist equipped with social interaction skills. DEVI's human-robot interface consists of an intelligence core and a hardware layer with 5 degrees-of-freedom. It has the capability to detect and recognize faces, register new people in the memory through conversation and maintain a database of people, greet them, respond to basic questions using an NLP platform and show directions with hands. It is a low cost solution, due to its low resource consumption and open-source core.\par

Instead of continuously executing the face recognition algorithm, we are using a hardware level proximity map to trigger the execution of face recognition in order to reduce CPU load and power consumption, which can be considered as a novel feature.
Furthermore, other notable features are the ability to recognize new people, collect their information via conversation and maintain a dynamic database of users, which removes the need for having a pre-trained fixed dataset for face recognition.\par

\subsection{Future Work}
Future developments of DEVI are planned in the aspects of adding more degrees-of-freedom to allow natural movements, deploy improved noise-cancelling mechanisms, decrease the latency of speech queries and adding additional actuators, in order to mobilize DEVI robot. 
Furthermore, face recognition system can be integrated with a data analytic system which could hold valuable information about people known to the system such as their availability at the premises, last arrival details. DEVI's core could be connected with other sensor systems to incorporate additional functionalities like directing people to the reception area. Thus, the open-source intelligence core of the DEVI system has the potential to become a solution to the increasing demand for socially assistive robots and socially interactive robots. 


{\small
\bibliography{devi}

\begin{thebibliography}{10}
\providecommand{\url}[1]{#1}
\csname url@samestyle\endcsname
\providecommand{\newblock}{\relax}
\providecommand{\bibinfo}[2]{#2}
\providecommand{\BIBentrySTDinterwordspacing}{\spaceskip=0pt\relax}
\providecommand{\BIBentryALTinterwordstretchfactor}{4}
\providecommand{\BIBentryALTinterwordspacing}{\spaceskip=\fontdimen2\font plus
\BIBentryALTinterwordstretchfactor\fontdimen3\font minus
  \fontdimen4\font\relax}
\providecommand{\BIBforeignlanguage}[2]{{%
\expandafter\ifx\csname l@#1\endcsname\relax
\typeout{** WARNING: IEEEtran.bst: No hyphenation pattern has been}%
\typeout{** loaded for the language `#1'. Using the pattern for}%
\typeout{** the default language instead.}%
\else
\language=\csname l@#1\endcsname
\fi
#2}}
\providecommand{\BIBdecl}{\relax}
\BIBdecl

\bibitem{feil2005defining}
D.~Feil-Seifer and M.~J. Mataric, ``Defining socially assistive robotics,'' in
  \emph{9th International Conference on Rehabilitation Robotics, 2005. ICORR
  2005.}\hskip 1em plus 0.5em minus 0.4em\relax IEEE, 2005, pp. 465--468.

\bibitem{fong2003survey}
T.~Fong, I.~Nourbakhsh, and K.~Dautenhahn, ``A survey of socially interactive
  robots,'' \emph{Robotics and autonomous systems}, vol.~42, no. 3-4, pp.
  143--166, 2003.

\bibitem{weller2017meet}
C.~Weller, ``Meet the first-ever robot citizen—a humanoid named sophia that
  once said it would’destroy humans’,'' \emph{Business Insider Nordic.
  Haettu}, vol.~30, p. 2018, 2017.

\bibitem{lafaye2014linear}
J.~Lafaye, D.~Gouaillier, and P.-B. Wieber, ``Linear model predictive control
  of the locomotion of pepper, a humanoid robot with omnidirectional wheels,''
  in \emph{2014 IEEE-RAS International Conference on Humanoid Robots}.\hskip
  1em plus 0.5em minus 0.4em\relax IEEE, 2014, pp. 336--341.

\bibitem{pepperprice}
G.~Robots, ``Pepper for business edition humanoid robot 2 years warranty,''
  \url{https://www.generationrobots.com/en/402422-pepper-for-business-edition-humanoid-robot-2-years-warranty.html},
  [Online; accessed 16-May-2019].

\bibitem{gershgorn2017inside}
D.~Gershgorn, ``Inside the mechanical brain of the world’s first robot
  citizen,'' \emph{QZ. Retrieved}, vol.~13, 2017.

\bibitem{aiko}
\BIBentryALTinterwordspacing
``{Humanoid robot starts work at Japanese department store}.'' [Online].
  Available:
  \url{www.reuters.com/article/us-japan-robot-store-idUSKBN0NB1OZ20150420}
\BIBentrySTDinterwordspacing

\bibitem{thalmann2017nadine}
N.~M. Thalmann, L.~Tian, and F.~Yao, ``Nadine: A social robot that can localize
  objects and grasp them in a human way,'' in \emph{Frontiers in Electronic
  Technologies}.\hskip 1em plus 0.5em minus 0.4em\relax Springer, 2017, pp.
  1--23.

\bibitem{c7}
\BIBentryALTinterwordspacing
``Vl53l0x tof sensor.'' [Online]. Available:
  \url{www.st.com/en/imaging-and-photonics-solutions/vl53l0x.html}
\BIBentrySTDinterwordspacing

\bibitem{c8}
\BIBentryALTinterwordspacing
``Cm785hb servo gearbox.'' [Online]. Available:
  \url{www.servocity.com/cm-785hb-servo-gearbox}
\BIBentrySTDinterwordspacing

\bibitem{c9}
\BIBentryALTinterwordspacing
``Springrc sr811 motor.'' [Online]. Available:
  \url{www.springrc.com/en/pd.jsp?id=81}
\BIBentrySTDinterwordspacing

\bibitem{c10}
\BIBentryALTinterwordspacing
``Micro maestro servo controller.'' [Online]. Available:
  \url{www.pololu.com/docs/0J40}
\BIBentrySTDinterwordspacing

\bibitem{c19}
TowerPro, ``Towerpro mg945 servo,''
  \url{http://www.towerpro.com.tw/product/mg945/}, [Online; accessed
  17-March-2019].

\bibitem{c11}
\BIBentryALTinterwordspacing
``Mongodb library.'' [Online]. Available: \url{www.mongodb.com/}
\BIBentrySTDinterwordspacing

\bibitem{c12}
M.~Liang and X.~Hu, ``Recurrent convolutional neural network for object
  recognition,'' in \emph{IEEE Conference on Computer Vision and Pattern
  Recognition (CVPR)}, 2015.

\bibitem{c13}
H.~et~al., ``Deep residual learning for image recognition,'' in \emph{IEEE
  Conference on Computer Vision and Pattern Recognition (CVPR)}, 2016.

\bibitem{c14}
\BIBentryALTinterwordspacing
``Dlib library.'' [Online]. Available: \url{http://dlib.net/}
\BIBentrySTDinterwordspacing

\bibitem{mic}
SeeedStudio, ``Respeaker mic array v2.0,''
  \url{https://media.digikey.com/pdf/Data%20Sheets/Seeed%20Technology/107990053_Br.pdf},
  [Online; accessed 17-March-2019].

\bibitem{c15}
\BIBentryALTinterwordspacing
``Dialogflow api.'' [Online]. Available: \url{https://dialogflow.com}
\BIBentrySTDinterwordspacing

\bibitem{manaswi2018speech}
N.~K. Manaswi, ``Speech to text and vice versa,'' in \emph{Deep Learning with
  Applications Using Python}.\hskip 1em plus 0.5em minus 0.4em\relax Springer,
  2018, pp. 127--144.

\bibitem{lundh1999introduction}
F.~Lundh, ``An introduction to tkinter,'' \emph{URL: www. pythonware.
  com/library/tkinter/introduction/index. htm}, 1999.

\bibitem{forrest2017essential}
C.~Forrest, ``Essential home wants to be ‘bridge’between amazon alexa,
  apple's siri, and google assistant,'' \emph{TechRepublic, May}, vol.~31,
  p.~9, 2017.

\bibitem{youtube}
\BIBentryALTinterwordspacing
R.~Karunasena. Devi: Human-robot interface for interactive receptionist
  systems. Youtube. [Online]. Available: \url{https://youtu.be/hvHou1GHkTc}
\BIBentrySTDinterwordspacing

\bibitem{video}
\BIBentryALTinterwordspacing
Karunasena. Ieee arm 2019: Devi. Youtube. [Online]. Available:
  \url{https://youtu.be/LRjZnNv2PPU}
\BIBentrySTDinterwordspacing

\bibitem{devigit}
``Devi: Open-source human-robot interface for interactive receptionist
  systems,'' \url{https://github.com/RameshaKaru/devi}, 2019.

\bibitem{c20}
U.~o.~M. Computer Vision~Laboratory, Computer Science~Department, ``Labeled
  faces in the wild,'' \url{http://vis-www.cs.umass.edu/lfw/}, [Online;
  accessed 17-March-2019].

\end{thebibliography}
\bibliographystyle{IEEEtran}
}

\end{document}